\documentclass[aps,pre,twocolumn,superscriptaddress]{revtex4-2}

\usepackage[utf8]{inputenc}
\usepackage[T1]{fontenc}
\usepackage{graphicx}
\usepackage{amsmath,amsthm}
\usepackage{amssymb}
\usepackage{nicefrac}
\usepackage{tikz}
\usepackage{tcolorbox}
\usepackage{hyperref} % hyperref should be loaded late
\usepackage{cleveref}
\usepackage{subcaption}
\usepackage{orcidlink}
\usetikzlibrary{positioning,arrows.meta,fit,calc}

\begin{document}

% --- Document Information ---
\title{Fast Bayesian equipment condition monitoring 
via simulation based inference: applications to heat exchanger health}

\author{Peter Collett}
\affiliation{Cognite AS, Oslo, Norway}

\author{Alexander Johannes Stasik,\orcidlink{0000-0003-1646-2472}}
\email{alexander.stasik@sintef.no} 

\affiliation{Department of Data Science, Norwegian University of Life Sciences, \AA s, Norway}
\affiliation{Department for Mathematics and Cybernetics, SINTEF Digital, Oslo, Norway}

\author{Simone Casolo\orcidlink{0000-0003-2945-8931}
}
\email{simone.casolo@cognite.com} 
\affiliation{Cognite AS, Oslo, Norway}

\author{Signe Riemer-Sørensen\orcidlink{0000-0002-5308-7651}
}
\affiliation{Department for Mathematics and Cybernetics, SINTEF Digital, Oslo, Norway}

\date{\today}

\begin{abstract}
Accurate condition monitoring of industrial equipment requires inferring latent degradation parameters from indirect sensor measurements under uncertainty. While traditional Bayesian methods like Markov Chain Monte Carlo (MCMC) provide rigorous uncertainty quantification, their heavy computational bottlenecks render them impractical for real-time process control. To overcome this limitation, we propose an AI-driven framework utilizing Simulation-Based Inference (SBI) powered by amortized neural posterior estimation to diagnose complex failure modes in heat exchangers. By training neural density estimators on a simulated dataset, our approach learns a direct, likelihood-free mapping from thermal-fluid observations to the full posterior distribution of degradation parameters. We benchmark this framework against an MCMC baseline across various synthetic fouling and leakage scenarios, including challenging low-probability, sparse-event failures. The results show that SBI achieves comparable diagnostic accuracy and reliable uncertainty quantification, while accelerating inference time by a factor of
82$\times$ compared to traditional sampling. The amortized nature of the neural network enables near-instantaneous inference, establishing SBI as a highly scalable, real-time alternative for probabilistic fault diagnosis and digital twin realization in complex engineering systems.
\end{abstract}

\maketitle


\section{Introduction}
The operational integrity and thermal efficiency of complex industrial systems are paramount to the economic performance and safety of modern process plants, driving a paradigm shift toward advanced predictive maintenance (PdM) and condition monitoring \cite{AHMEDMURTAZA2024102935, villa2025application}.
A foundational challenge in industrial asset management is that 
critical process or health parameters 
such as efficiency or degradation coefficients, internal component wear, 
or leak rates often
cannot be measured directly with standard instrumentation. 
Instead, these latent variables can be inferred from observable sensor streams by leveraging 
mathematical models that simulate the physical behavior of the equipment, often by a costly manual trial-and-error process. 
While this model-based
approach is broadly applicable to a wide range of industrial equipment relying on first-principles 
or empirical simulations, it is particularly vital for systems where internal states are 
inaccessible during operation \cite{ruosen_qi_9a04489d, atma_sahu_77b6a77b}. 
Within this context, direct Bayesian inference provides a rigorous framework for estimating these latent parameters from sensor data, enabling the uncertainty-aware calibration of physical simulators and digital twins
to accurately reflect the true equipment or process state.\\
Heat exchangers serve as a prototypical 
example of industrial equipment where health parameters must be indirectly determined. 
Machine learning and deep learning methods have been extensively studied in the last decade for condition monitoring of heat exchangers failures \cite{Fouling_deeplearnign3, SUNDAR2020120112, WU2023119285, fouling_DeepLearning1, fouling_DeepLearning2}, with recent momentum heavily favoring time-series networks like LSTMs, explainable gradient boosting (XGBoost), neural networks \cite{EAAI_HeatTransfer_2025} and hybrid ML ensembles \cite{HOU2025108809, HOU2026111129, pr13010219, HOSSEINI20228767, BERCE2025126954, ABOULKHAIL2026104565}.
For these 
units, essential variables such as the fouling 
resistance ($R_f$), the effective heat transfer area 
($A$), and internal mass-loss fractions are unobservable directly. Instead, these parameters are  
estimated from observable data, such as inlet and outlet temperatures or fluid mass flow rates, 
utilizing thermal-fluid simulation tools to bridge the gap between sensor readings and the 
underlying equipment state \cite{en16062812,ETMINAN2025100022}. 
Bayesian inference provides a rigorous mathematical framework for 
resolving this inverse problem by treating these unobservable parameters as random variables \cite{gelman1995bayesian}. 
This allows for the quantification of uncertainty through a posterior probability distribution, which is essential for risk-aware decision-making and predicting Remaining Useful Life (RUL) \cite{MORADI2022108433, cholette2019degradation, BARDEENIZ2025111250}.
However, the practical application of traditional Bayesian tools, such as Markov Chain Monte Carlo (MCMC) \cite{galin_l__jones_610ba213, XIAO2026111718} or Sequential Monte Carlo (Particle Filtering) \cite{ZIO2011403}, is severely constrained by computational overhead when scaled to complex systems.
To ensure convergence, MCMC samplers typically require thousands of 
iterative evaluations of the underlying physical simulation model for every single inference call. This computational bottleneck renders MCMC impractical for online monitoring 
scenarios where rapid, high-frequency diagnostics are required for real-time process control. 
To address these limitations, modern Artificial Intelligence (AI) paradigms, including normalizing flows and neural-networked variational inference \cite{NAZEMZADEH2026100030, MO2025111337, DASGUPTA2024109729}, and specifically Simulation-Based Inference (SBI) \cite{cranmer2020frontier,deistler2025simulationbasedinferencepracticalguide}, have emerged as scalable alternatives to bypass high-dimensional Bayesian inverse bottlenecks.
The primary advantage of SBI is its 
ability to perform likelihood-free inference by leveraging the forward process of the simulator 
itself. By generating a comprehensive dataset pairing parameter inputs and simulation outputs in a one-time offline phase, SBI employs neural density estimators to learn the direct mapping from observed data to the full distribution of the posterior, likelihood or likelihood ratio \cite{cranmer2020frontier}. Once the neural network is trained, the computational burden is effectively amortized; subsequent inference calls for new sensor measurements require only seconds, allowing the method to scale across complex industrial systems and multiple assets simultaneously \cite{_lvaro_tejero_cantero_3be1efca}. Simultaneously, the neural network can be easily updated with additional simulations if observations drift beyond the range covered by the originally sampled pairs of parameter input and simulation output. While SBI is established as a scientific method \cite{cranmer2020frontier}, the literature lacks examples of industrial applications and performance on such systems.
In this study, we present a probabilistic framework for the 
automated diagnosis of failure modes in a shell-and-tube heat exchanger, focusing specifically on fouling and leakage scenarios. 
We establish stochastic models for two primary failure mechanisms, progressive fouling and internal leakage, governed by latent parameters to be determined in a Bayesian fashion. A systematic comparison is conducted between traditional MCMC sampling and amortized SBI, specifically utilizing sequential Neural Posterior Estimation (NPE, \cite{papamakarios2018fastepsilonfreeinferencesimulation}) to identify the onset and evolution of these failures. Our results demonstrate that the SBI approach is orders of magnitude faster than MCMC, providing near-instantaneous posterior characterization without sacrificing diagnostic depth. By highlighting this significant gain in computational efficiency, this work establishes a scalable workflow for deploying high-fidelity Bayesian condition monitoring in real-time thermal-fluid applications. Furthermore, we remark how this approach is transferrable to the condition monitoring of other multi-parameter industrial processes and equipment making it exceptionally well-suited for legacy systems or "black-box" simulators where the underlying governing equations are inaccessible \cite{pmlr-v89-papamakarios19a}. 
Alternative methods include learning fast approximations for the simulator \cite{HE2025113759} to speed up classical MCMC implementation, replacing legacy simulators with their differentiable counterparts \cite{degrave2018differentiablephysicsenginedeep}, or leveraging Physics-Informed Neural Networks (PINNs) to embed governing physical constraints directly into the learning architecture \cite{Majumdar03062025, shi2025hybrid}.

\section{Methods} \label{sec:methods}
As physical basis for our diagnostic model, we study a deterministic heat exchanger model (Sec. \ref{sec:deterministicHX}) with stochastic failures (Sec. \ref{sec:stochasticHX}). 

\subsection{Deterministic Heat Exchanger Model} \label{sec:deterministicHX}

\begin{figure}[t]
    \centering
    \begin{tikzpicture}[scale=0.82, thick, >=stealth]
        % Define colors for temperature gradients
        \colorlet{hotin}{red!95!black}
        \colorlet{hotout}{orange!85!yellow}
        \colorlet{coldin}{blue!95!black}
        \colorlet{coldout}{cyan!90!white}

        % --- Heat Exchanger Shell ---
        \draw[fill=gray!5, draw=gray!60, line width=1pt] (0,-0.1) rectangle (6,2.5);
        \node[gray!80, font=\sffamily\bfseries] at (3,2.8) {Shell-and-Tube Heat Exchanger};

        % --- Internal Baffles ---
        \foreach \x in {1.2, 2.4, 3.6, 4.8} {
            \draw[gray!40, line width=2pt] (\x, -0.1) -- (\x, 1.6);
        }
        \foreach \x in {0.6, 1.8, 3.0, 4.2, 5.4} {
            \draw[gray!40, line width=2pt] (\x, 2.5) -- (\x, 0.8);
        }

        % --- Hot Fluid Path (Tube Side) ---
        \shade[left color=hotin, right color=hotout]
        (-1,1.72) rectangle (6.8,1.88);
        \draw[line width=3.5pt, hotout, ->] (6.8,1.8) -- (7.2,1.8);
                
        % Labels for Hot Fluid
        \node[hotin, left] at (-1.1, 1.8) {\textbf{$T_{\mathrm{hot,in}}$}};
        \node[hotout, right] at (7.1, 1.8) {\textbf{$T_{\mathrm{hot,out}}$}};
        \node[hotin!80!black, above] at (3, 1.9) {$\dot{m}_{\mathrm{hot}}, c_{p,\mathrm{hot}}$};

        % --- Cold Fluid Path (Shell Side) ---
        \shade[left color=coldout, right color=coldin]
        (-0.8,0.47) rectangle (7,0.63);
        \draw[line width=3.5pt, coldout, ->] (-0.6,0.55) -- (-1.0,0.55);
        
        % Labels for Cold Fluid
        \node[coldin, right] at (7.1, 0.6) {\textbf{$T_{\mathrm{cold,in}}$}};
        \node[coldout, left] at (-1.1, 0.4) {\textbf{$T_{\mathrm{cold,out}}$}};
        \node[coldin!80!black, below] at (3, 0.485) {$\dot{m}_{\mathrm{cold}}, c_{p,\mathrm{cold}}$};

        % --- Heat Transfer Variables ---
        % Central UA indicator
        \draw[<->, black!70, dashed, line width=0.8pt] (3, 0.8) -- (3, 1.6) 
            node[midway, fill=gray!5, inner sep=1pt] {$UA$};
        
        % Heat Transfer Arrow
        \draw[->, orange!90!black, ultra thick] (4, 1.6) -- (4, 0.9) 
            node[midway, right] {$Q$};
    \end{tikzpicture}
    \caption{Schematic of a counterflow heat exchanger with key variables. The hot fluid (red) enters at $T_{\mathrm{hot,in}}$ with mass flow rate $\dot{m}_{\mathrm{hot}}$ and specific heat $c_{p,\mathrm{hot}}$, and exits at $T_{\mathrm{hot,out}}$. The cold fluid (blue) enters at $T_{\mathrm{cold,in}}$ with $\dot{m}_{\mathrm{cold}}$ and $c_{p,\mathrm{cold}}$, and exits at $T_{\mathrm{cold,out}}$. Heat, $Q$, is transferred across the exchanger wall with overall conductance $UA$.}
    \label{fig:hex_sketch}
\end{figure}
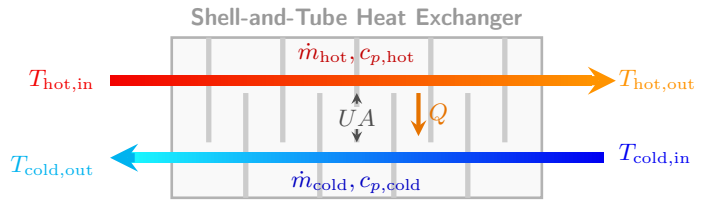
For the deterministic model, we consider a steady-state counterflow heat exchanger as illustrated in Fig. \ref{fig:hex_sketch}. The heat capacity rates for the hot and cold fluid streams are defined by their respective mass flow rates $\dot{m}$ and specific heat capacities $c_p$:
\begin{align}
    C_\mathrm{hot} &= \dot{m}_\mathrm{hot} c_{p,\mathrm{hot}}\, , & C_\mathrm{cold} &= \dot{m}_\mathrm{cold} c_{p,\mathrm{cold}} \, .
\end{align}
For a counterflow heat exchanger, the energy (heat)
balances are given by
\begin{align}
Q &= \dot{m}_{\mathrm{hot}} \, c_{p,\mathrm{hot}} (T_{\mathrm{hot,in}} - T_{\mathrm{hot,out}})\, , \\ 
Q &= \dot{m}_{\mathrm{cold}} \, c_{p,\mathrm{cold}} (T_{\mathrm{cold,out}} - T_{\mathrm{cold,in}}) \, ,
\end{align}
as well as the heat transfer equation,
\begin{equation}
Q = UA \cdot \Delta T_{\mathrm{LM}}, 
\end{equation}
where $\Delta T_{\mathrm{LM}}$ denotes the log-mean temperature 
difference (LMTD) between the two streams, defined as
\begin{equation}
\Delta T_{\mathrm{LM}} = 
\frac{(T_{\mathrm{hot,in}} - T_{\mathrm{cold,out}}) - (T_{\mathrm{hot,out}} - T_{\mathrm{cold,in}})}
{\ln \left( \frac{T_{\mathrm{hot,in}} - T_{\mathrm{cold,out}}}{T_{\mathrm{hot,out}} - T_{\mathrm{cold,in}}} \right)} \, .
\end{equation}
While the heat transfer rate $Q$ is traditionally solved
using the LMTD method, 
the nonlinear nature of the LMTD requires iterative root-finding 
for unknown outlet temperatures. To facilitate efficient Bayesian 
inference, we utilize the effectiveness-NTU ($\epsilon$-NTU)
method \cite{bergman2011fundamentals}, which provides an explicit 
analytical solution. The heat transfer is expressed as:
\begin{equation}
    Q = \epsilon C_{\min} (T_{\mathrm{hot,in}} - T_{\mathrm{cold,in}}) \, ,
\end{equation}
where $C_{\min} = \min(C_\mathrm{hot}, C_\mathrm{cold})$. The heat exchanger effectiveness, $\epsilon$, represents the ratio of actual heat transfer to the maximum thermodynamic limit. Here, $\epsilon$ is determined by the Number of Transfer Units ($\text{NTU}$) and the capacity rate ratio $r$:
\begin{equation}
    \text{NTU} = \frac{UA}{C_{\min}} \, , \quad r = \frac{C_{\min}}{C_{\max}} \, ,
\end{equation}
\begin{equation}
    \epsilon = \frac{1 - \exp[-\text{NTU}(1 - r)]}{1 - r \exp[-\text{NTU}(1 - r)]} \, .
\end{equation}
The product of the overall heat transfer coefficient $U$ and the exchange area $A$ is denoted as $UA$, which characterizes the heat transfer capability of the equipment.
The outlet temperatures are then directly recovered from the
energy balance:
\begin{align*}
    T_{\mathrm{hot,out}} &= T_{\mathrm{hot,in}} - \frac{Q}{C_\mathrm{hot}} \\ 
    T_{\mathrm{cold,out}} &= T_{\mathrm{cold,in}} + \frac{Q}{C_\mathrm{cold}}. 
\end{align*}

\subsection{Stochastic Modeling of Failure Mechanisms} \label{sec:stochasticHX}
For each of the failure mechanisms object of this study, degradation is modeled by introducing a stochastic time-dependency into the parameters of the deterministic model in Sec. \ref{sec:deterministicHX}, aligning with recent reliability frameworks for stochastic degrading devices \cite{ZHANG2024110223}.
These stochastic formulations are selected not because they necessarily provide a more accurate physical representation of degradation than existing deterministic models, but rather as tunable, multi-parameter frameworks for failure evolution. By adjusting the event frequency and the severity scales of the failure mechanisms, we can generate a wide spectrum of failure scenarios. This flexibility allows for the systematic creation of datasets that vary in their level of difficulty for anomaly detection and parameter identification. 
Consequently, the model serves as a rigorous testbed for evaluating the robustness of the Bayesian inference framework across scenarios ranging from subtle, continuous degradation to sporadic, high-impact failure events. In general, we assume failures initiate at an unknown changepoint $\tau$. To maintain differentiability for gradient-based inference, we utilize a logistic sigmoid transition $S(t)$ to model the induction period:
\begin{equation}
    S(t) = \frac{1}{1 + \exp(-k(t - \tau))} \, , \label{eq:tau}
\end{equation}
where $k$ is the transition sharpness. Two primary failure mechanisms are considered:

\textbf{1. Tube Fouling.}
\begin{figure}
    \centering
    \includegraphics[width=0.95\linewidth]{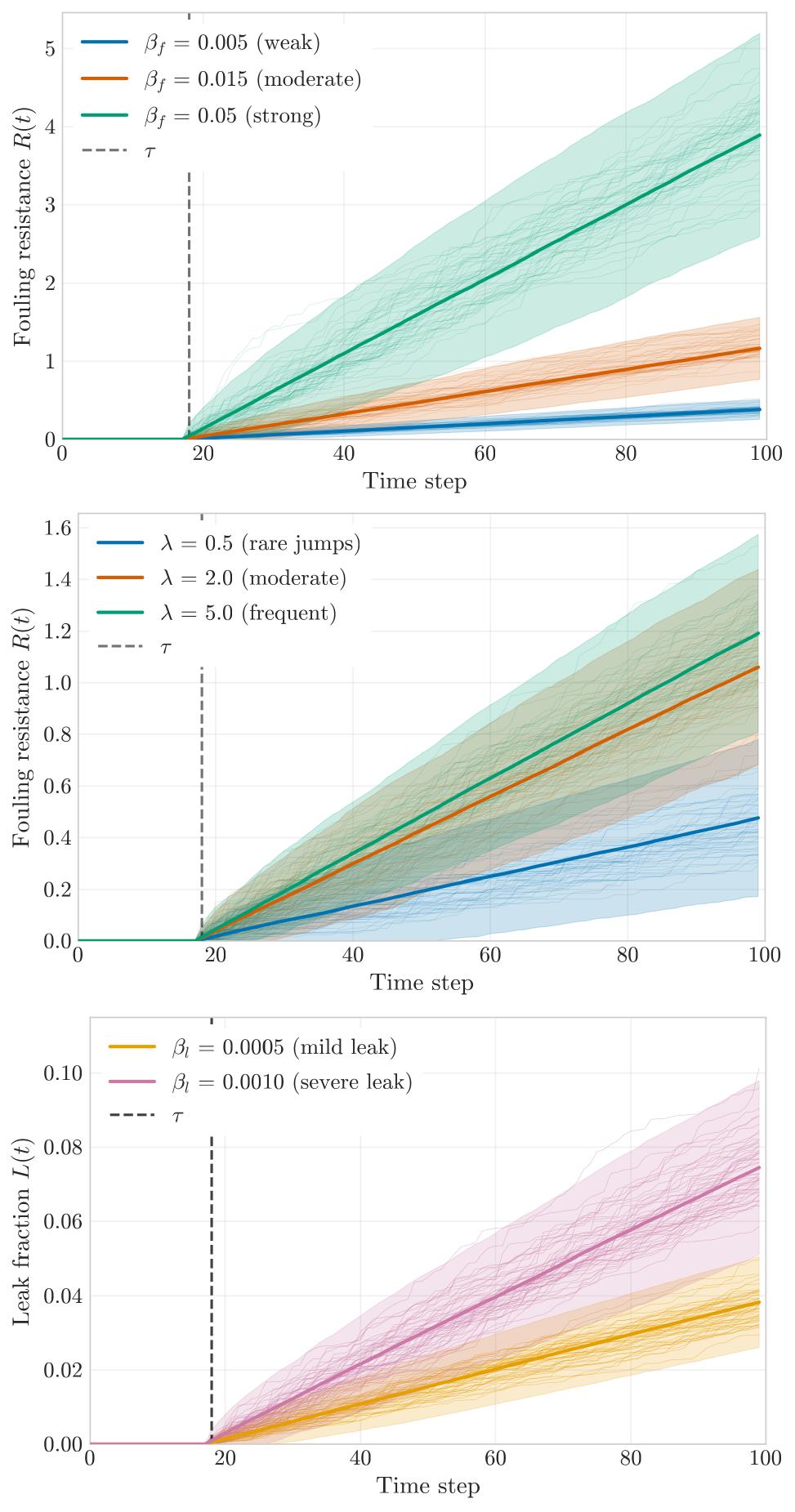}

    \caption{Fouling and Leakage evolution in time with the effect of the failure parameters ($\lambda, \beta_f, \beta_l$) over
    the trajectories at a fixed failure onset time $\tau$.}
    \label{fig:fouling_model}
\end{figure}
In this work, grounded in recent analyses of physical deposition mechanisms \cite{WANG2026128183}, fouling is modeled as a reduction in the overall conductance $UA$ via a non-negative, dimensionless fouling factor $R(t)$:
\begin{equation}
    UA(t) = \frac{UA_{\mathrm{clean}}}{1 + R(t)} \, .
\end{equation}
To capture the sporadic, burst-like nature of industrial scaling, $R(t)$ is modelled as a discretized and relaxed Compound Poisson Process \cite{tankov2003financial}. The total fouling at time $t$ is the accumulation of stochastic increments:
\begin{equation}
    R(t) = \sum_{i=1}^{t} \Delta R_i, \qquad \Delta R_i = S(i) \cdot \mathbb{I}_i \cdot J_i \cdot g_f \, ,
\end{equation}
where $g_f \in \{0,1\}$ is a binary fouling indicator determined by the sampled failure mode $z$. Specifically, $g_f=1$ when the selected mode includes fouling and $g_f=0$ otherwise. The term $S(i)$ represents a steep sigmoid function centered at the random failure time $\tau$, acting as a temporal gate that activates the process for $i \geq \tau$.
The stochastic occurrence of a jump is governed by the relaxed gating variable $\mathbb{I}_i$. At each time step, a jump probability is calculated from the Poisson arrival rate $\lambda$ as $P_{\text{jump}} = 1 - \exp(-\lambda)$. The operator is defined by passing a latent uniform sample $u_i \sim \text{Uniform}(0,1)$ through a steep sigmoid function:
\begin{equation}
    \mathbb{I}_i = \frac{1}{1 + \exp(-k (P_{\text{jump}} - u_i))} \, .
\end{equation}
The magnitude of the potential jump, $J_i$, is drawn from an exponential distribution. To ensure a mean jump size of $\beta_f$, the distribution is parameterized by the rate $1/\beta_f$:
\begin{equation}
    J_i \sim \text{Exp}\left( \frac{1}{\beta_f} \right), \quad \text{where } \mathbb{E}[J_i] = \beta_f \, . \label{eq:beta_f}
\end{equation}
Overall, the fouling process depends on the arrival rate $\lambda$ and the strength parameter $\beta_f$. As $\lambda$ increases, the frequency of stochastic events grows, while a higher $\beta_f$ increases the average magnitude of each individual jump.\\
The expected growth of the fouling factor can be analyzed via the linearity of expectation. For the post-induction phase ($i > \tau$), where the temporal gate $S(i) \approx 1$, the expected increment is the product of the expected jump probability and the expected jump magnitude:
\begin{equation}
    \mathbb{E}[\Delta R_i] = \mathbb{E}[\mathbb{I}_i] \cdot \mathbb{E}[J_i] \cdot g_f = (1 - e^{-\lambda}) \cdot \beta_f \cdot g_f \, .
\end{equation}
For $t > \tau$, the total accumulated fouling is the sum of these independent increments over the active duration of the failure. Thus, by Wald identity, 
the expectation of $R(t)$ is linear in time:
\begin{equation}
    \mathbb{E}[R(t)] \approx \left[(1 - e^{-\lambda}) \, \beta_f \, g_f\right] (t - \tau) \, .
\end{equation}
compatibly with the well known Ebert-Panchal model \cite{Ebert-Panchal}.
This derivation shows that the process is linear in time relative to the random failure point $\tau$. This formulation is consistent with the first-order behavior of deterministic fouling laws while retaining the random-event structure necessary to represent the intermittent and unpredictable nature of industrial fouling.

\textbf{2. Leakage with Fluid Loss.} 
Internal leakage is modeled as a diversion of fluid from the primary hot-stream flow path, representing a shell-side bypass or a loss of integrity in the tube-to-header seals. We define a time-dependent leak fraction $L(t) \in [0, 1)$ that reduces the effective mass flow rate participating in the heat exchange:
\begin{equation}
    \dot{m}_{\mathrm{hot,out}}(t) = \dot{m}_{\mathrm{hot,in}} (1 - L(t)) \, .
\end{equation}
This reduction in the hot-side thermal mass decreases the heat capacity rate $C_\mathrm{hot}$, altering the capacity rate ratio $r$. This results in a simultaneous reduction in outlet mass flow and a decrease in $T_{\mathrm{hot,out}}$.\\
In our probabilistic framework, the evolution of the leak fraction $L(t)$ is modeled as a continuous stochastic growth process. Following the changepoint $\tau$, the cumulative degradation is driven by exponential increments:
\begin{equation} 
    L(t) = L_{\max}\left(1 - \exp\left[-\sum_{i=1}^{t} S(i)\cdot g_l \cdot \Delta L_i\right]\right) \, , \label{eqn:leak}
\end{equation}
where $L_{\max} = 0.95$ represents the physical saturation of the leak and $g_l \in \{0,1\}$ is a binary leakage indicator determined by the sampled failure mode $z$. The increments $\Delta L_i$ are sampled from an exponential distribution $\Delta L_i \sim \exp{(1/\beta_l)}$, where the scale parameter $\beta_l$ determines the growth rate of the failure.

\subsection{Bayesian Inference and Probabilistic Modeling}

Bayesian inference offers a rigorous mathematical framework for quantifying uncertainty in model parameters $\theta$ given observed data $\mathbf{y}$. 
In the case of industrial simulations, it solves the inverse problem of determining 
the distribution $p(\theta \mid \mathbf{y})$ representing the probability that 
a set of parameters $\theta$ produce the observed sensor data $\mathbf{y}$.
The cornerstone of the Bayesian approach is the posterior distribution,
\begin{equation}
    p(\theta \mid \mathbf{y}) = \frac{p(\mathbf{y} \mid \theta)\,p(\theta)}{p(\mathbf{y})},
\end{equation}
where $p(\theta)$ is the prior, $p(\mathbf{y} \mid \theta)$ is the likelihood, and $p(\mathbf{y}) = \int p(\mathbf{y} \mid \theta)\,p(\theta)\,d\theta$ is the model evidence. In the context of heat exchanger modeling, $\theta$ may include parameters such as the effective heat transfer coefficient, fouling factor, mass flow rates, and failure mode indicators. The likelihood is typically constructed based on a well-defined physical model such as the one in Sec.~\ref{sec:stochasticHX}, incorporating measurement noise and potential process disturbances:
\begin{equation}
    \mathbf{y}_i = f(\theta, \mathbf{x}_i) + \epsilon_i, \qquad \epsilon_i \sim \mathcal{N}(0, \sigma^2),
\end{equation}
where $f(\cdot)$ denotes the deterministic model mapping, $\mathbf{x}_i$ are covariates or known inputs, and $\epsilon_i$ are independent, normally distributed errors.

Posterior inference enables both point and interval estimation, as well as predictive uncertainty. The predictive distribution for new data $\mathbf{y}^*$ given new input $\mathbf{x}^*$ is
\begin{equation}
    p(\mathbf{y}^* \mid \mathbf{y}) = \int p(\mathbf{y}^* \mid \theta, \mathbf{x}^*)\,p(\theta \mid \mathbf{y})\,d\theta,
\end{equation}
fully accounting for parameter uncertainty. In most realistic settings, this integral cannot be evaluated analytically. Instead, posterior inference must rely on numerical methods such as Markov Chain Monte Carlo (MCMC, \cite{vanRavenzwaaij2018MCMC}) or variational inference \cite{Blei_2017}.
MCMC constructs a Markov chain whose stationary distribution is the posterior \(p(\theta \mid \mathbf{y})\), which enables asymptotically exact sampling-based inference for both continuous and discrete latent variables. In this work, we use Hamiltonian Monte Carlo through the No-U-Turn Sampler (NUTS) for continuous parameters \cite{NUTS}. For the discrete failure-mode formulation we combine NUTS with a Gibbs update over the categorical mode variable. MCMC methods provide a strong posterior benchmark, but becomes computationally expensive as it requires a new sampling procedure for every observation.

\subsection{Simulation-Based Inference (SBI)}
\begin{figure*}[t]
    \centering
    \includegraphics[width=1\textwidth]{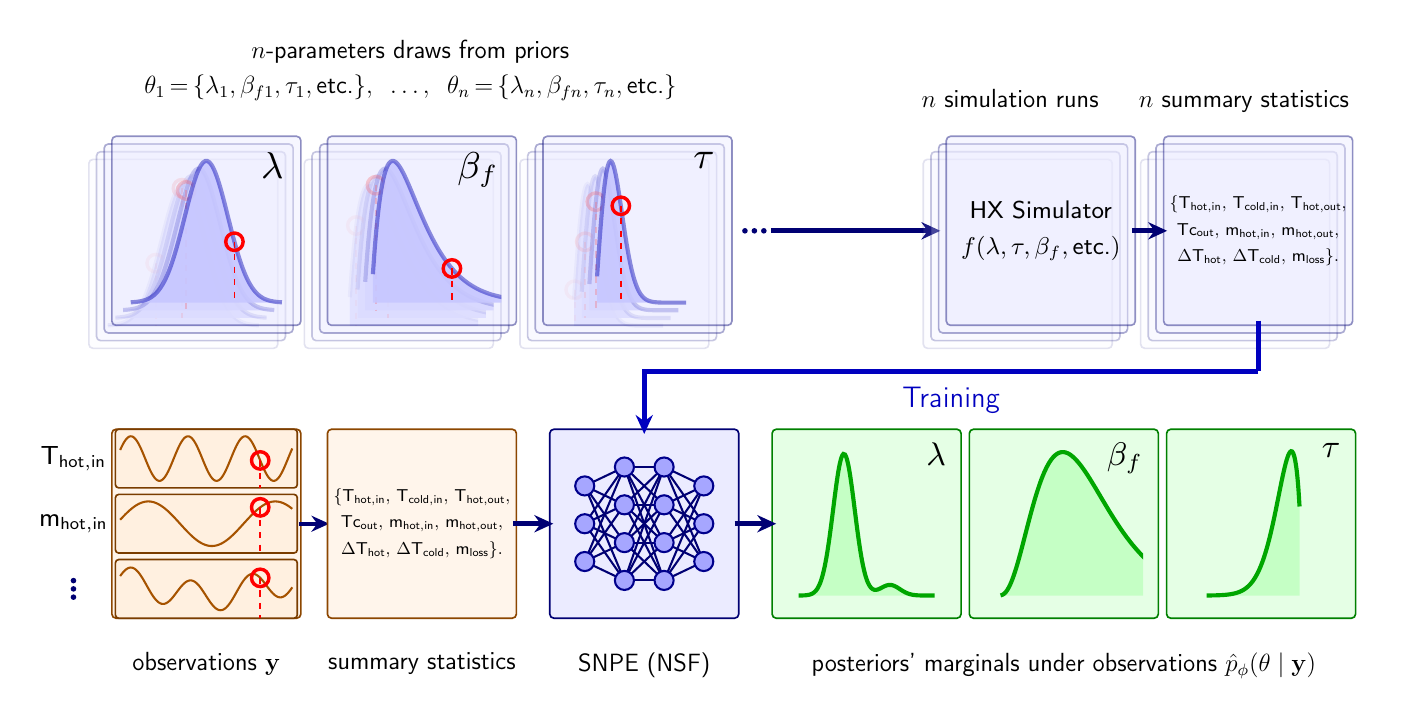}
    \caption{The applied SBI scheme. 
    Blue: training data (priors) and process, orange: input data, green: inferred output posteriors. 
    In this work $n$=50,000 simulations.} 
    \label{fig:SBI}
\end{figure*}
Simulation-Based Inference (SBI), also known as likelihood-free inference, is a family of techniques for Bayesian inference in models for which the likelihood function $p(\mathbf{y} \mid \theta)$ is analytically intractable or unavailable, but simulation of synthetic data $\mathbf{y}_\mathrm{sim} \sim p(\mathbf{y} \mid \theta)$ is feasible~\cite{cranmer2020frontier}. This is particularly relevant for nonlinear physical systems with complex failure dynamics such as the heat exchanger.\\
One classical SBI approach is Approximate Bayesian Computation (ABC)
\cite{ABC}, which defines an approximate posterior as
\begin{equation}
    p_\epsilon(\theta \mid \mathbf{y}) \propto p(\theta) \int \mathbb{I}\left(d(\mathbf{y}, \mathbf{y}_\mathrm{sim}) < \epsilon\right) p(\mathbf{y}_\mathrm{sim} \mid \theta) d\mathbf{y}_\mathrm{sim},
\end{equation}
where $d(\cdot,\cdot)$ is a distance metric and $\epsilon$ is a user-defined tolerance. More recent methods leverage neural density estimators to approximate either the posterior $p(\theta \mid \mathbf{y})$, the likelihood $p(\mathbf{y} \mid \theta)$, or the likelihood-to-evidence ratio, typically via simulated data sets \cite{papamakarios2018fastepsilonfreeinferencesimulation}:
\begin{equation}
    \hat{p}_\phi(\theta \mid \mathbf{y}) \approx p(\theta \mid \mathbf{y}),
\end{equation}
where $\hat{p}_\phi$ denotes the neural network approximation parameterized by $\phi$. These methods permit inference even in complex models for which classical Bayesian updating is not tractable, provided that forward simulation is possible. For recent advances and theoretical guarantees, see~\cite{cranmer2020frontier}.
We apply an amortized sequential neural posterior estimation (NPE) \cite{Greenberg2019AutomaticPT} as implemented in the \texttt{sbi toolkit} software package \cite{BoeltsDeistler_sbi_2025}. Details on the SBI approach can be found in Figure~\ref{fig:SBI} and Section~\ref{sec:exp}.

\subsection{Summary statistics}
In Bayesian inference such as MCMC and SBI, the data and models are compared on so-called summary statistics. For high-dimensional data, lower dimensional representations that preserve the relevant information about the parameters of interest enable comparison to simulated data. In particular, they are crucial when the system is complex and the likelihood function is intractable. The lower dimensional representation can be designed from knowledge or learned through an embedding neural network. In this study, we construct a set of summary statistics tailored to the heat exchanger measurements. Features are derived from the hot-side temperature difference \(T_\mathrm{hot,in}-T_\mathrm{hot,out}\), the cold-side temperature difference \(T_\mathrm{cold,out}-T_\mathrm{cold,in}\), the hot-side mass-flow loss \(\dot{m}_\mathrm{hot,in}-\dot{m}_\mathrm{hot,out}\), and the outlet temperatures \(T_\mathrm{hot,out}\) and \(T_\mathrm{cold,out}\). For each of these five signals, the mean, standard deviation, early-to-late change, dynamic range, and a linear trend measure are computed to yield a 25-dimensional summary statistic vector. These summaries are designed to capture the temporal signatures most informative for both failure-mode identification and estimation of the associated degradation parameters.

\subsection{Scoring and distance measures}
To compare performance of SBI against MCMC, we use metrics that assess both the similarity between inferred posteriors and their quality with respect to the ground-truth parameters. The Wasserstein distance between the SBI and MCMC posteriors quantifies the discrepancy between the amortized SBI posterior and the MCMC reference posterior. The Continuous Ranked Probability Score (CRPS) and credible-interval coverage are used to evaluate the quality of each inferred posterior. These metrics are evaluated separately for each continuous parameter of interest.\\
CRPS is a generalization of the Mean Absolute Error (MAE) for probabilistic predictions \cite{CRPS}. It is calculated by comparing the empirical cumulative distribution function (ECDF) of posterior samples with the step function CDF at the true value. Lower CRPS indicates a posterior that is sharper and better aligned with the ground-truth.\\
The Wasserstein distance is instead a similarity metric between two probability distributions. The convergence rate of empirical Wasserstein estimators typically behaves as $n^{-1/d}$, where $n$ is the number of samples and $d$ is the dimension, implying that standard empirical Wasserstein distance becomes a poor measure of similarity for high-dimensional data. It is also sensitive to extreme outliers. 

%Scipy implementation %https://docs.scipy.org/doc/scipy/reference/generated/scipy.stats.wasserstein_distance_nd.html
%Jenssen-Shannon is only one-dimensional
%\subsection{N-dimensional Kolmogorov-Smirnov Distance}
%https://arxiv.org/abs/2504.11299

\section{Experiments}
The aim is to evaluate the robustness of the Bayesian diagnostic framework by identifying the failure mode and determine the failure parameters: induction time $\tau$ (Eqn.~\ref{eq:tau}), fouling jump scale $\beta_f$ (Eqn.~\ref{eq:beta_f}), leakage growth $\beta_l$ (Eqn.~\ref{eqn:leak}) and fouling intensity $\lambda$. In the model setup, failure mode is represented by a discrete categorical variable \(z \in \{\text{none}, \text{fouling}, \text{leakage}, \text{both}\}\), rather than a continuous simplex Dirichlet prior over mode probabilities. This avoids the confounding present in the Dirichlet formulation, where the mode weights also scale degradation magnitude. As a result, the categorical formulation leads to clearer failure-mode identification.

\subsection{Test scenarios}
For evaluation purposes, we define five distinct failure scenarios representing a spectrum of industrial operational conditions. These scenarios are generated using the stochastic model described in Sec.~ \ref{sec:methods}, with specific parameters detailed in Table \ref{tab:scenario_parameters}. 
\begin{table}[t]
\centering
\small
\begin{tabular}{l l c c c c}
\hline
\textbf{Scenario} & \textbf{Failure} & \textbf{$\tau$} & 
\textbf{$\beta_f$} & \textbf{$\beta_l$} & \textbf{$\lambda$}\\
\hline
Weak Fouling   & Fouling & 18 & 0.005 & - & 5.0 \\
Batch SD       & Fouling & 18 & 0.030 & - & 0.5 \\
Boiler FW      & Fouling & 18 & 0.050 & - & 3.0 \\
Mild Leak      & Leakage & 18 & - & 0.0005 & - \\
Severe Leak    & Leakage & 18 & - & 0.0010 & - \\
\hline
\end{tabular}
\caption{Failure parameters for synthetic failure scenarios. $\tau$ denotes the failure changepoint (induction time); $\beta_f$ and $\beta_l$ represent the scale parameters for fouling jumps and leakage growth, respectively; and $\lambda$ defines the fouling arrival rate (intensity). 
%Italicized entries indicate inactive nuisance parameters that were fixed at their prior median values during data generation but do not influence the generated observations under the specified failure mode. A dash indicates that the parameter is not operationally defined for the scenario.
}
\label{tab:scenario_parameters}
\end{table}
The scenarios are designed to test the model's ability to distinguish between quasi-continuous degradation and sporadic, high-impact events:
\begin{description}
     \item[Weak Fouling (Scenario 1)] This represents a well-maintained system where fouling is slow and progresses in a quasi-continuous manner. The high arrival rate ($\lambda = 5.0$) combined with a low jump scale ($\beta_f = 0.005$) simulates frequent, minor deposition events that approximate a linear degradation trend. This is typical of systems with high fluid velocity or effective surface conditioning, presenting a challenge for early-stage anomaly detection due to the subtle signal-to-noise ratio of the degradation increments.
    \item[Batch Process Shutdown (Scenario 2)] This scenario simulates fouling that occurs primarily during stagnant periods or process shutdowns (Batch SD). The low arrival rate ($\lambda = 0.5$) and high jump scale ($\beta_f = 0.03$) result in rare but severe ``shocks'' to the heat transfer efficiency. The diagnosis of such low-probability, time-dependent rare events remains a recognized challenge in industrial fault detection \cite{Pourmir2025EAAI}, particularly under few-shot observability constraints where Bayesian uncertainty calibration is vital \cite{CHANG2025109980}.
    \item[Boiler Feedwater System (Scenario 3)] Representing aggressive scaling in systems with high mineral content (Boiler FW), this scenario utilizes both high frequency and high severity scales. The combination of $\lambda = 3.0$ and $\beta_f = 0.05$ creates a rapid loss of heat transfer capability. This scenario tests the model's ability to track high-gradient failure progression and its reliability in predicting the remaining useful life of the equipment.
\end{description}

The leakage scenarios represent progressive structural failures, such as tube pitting or gasket degradation. In both cases, the failure is characterized by a gradual loss of mass flow and a corresponding thermal fingerprint in the outlet temperatures.
\begin{description}
    \item[Mild Leak (Scenario 4)] A slow-onset failure where the leak fraction grows moderately ($\beta_l = 0.0005$). This scenario tests the sensitivity of the Bayesian framework to distinguish small mass-flow discrepancies from measurement noise.
    \item[Severe Leak (Scenario 5)] A rapid loss of system integrity ($\beta_l = 0.0010$). This scenario creates a significant and fast-evolving divergence between inlet and outlet mass flow, requiring the model to maintain stability while the heat capacity rate ratio $r$ shifts rapidly.
\end{description}
Finally, we also include a baseline no-failure case:
\begin{description}
    \item[No Failure (Scenario 6)] This scenario represents normal heat exchanger operation without fouling or leakage. It serves as a baseline for assessing whether the inference framework can correctly identify the absence of failure and avoid false-positive diagnoses.
\end{description}
For each scenario, an ensemble of 500 synthetic datasets was generated to ensure statistical significance in the evaluation of the inference methods. For scenario identification, each discrete failure mode is handled by predicting logits, which are transformed via a softmax function to obtain mode probabilities. At inference time, we draw posterior samples, assign each draw the label with the highest softmax probability, and thus obtain discrete samples of the latent realized trajectory, $z$, analogous to the MCMC output. A scenario is considered correctly identified if the most frequent posterior label matches the true simulated label. In addition to the failure mode status label, we also infer the failure parameters $\{\tau, \lambda, \beta_f, \beta_l \}$ directly from the data and independently of the failure category.

\subsection{Priors}
\begin{figure}[t]
    \centering
    \includegraphics[width=0.8\linewidth]{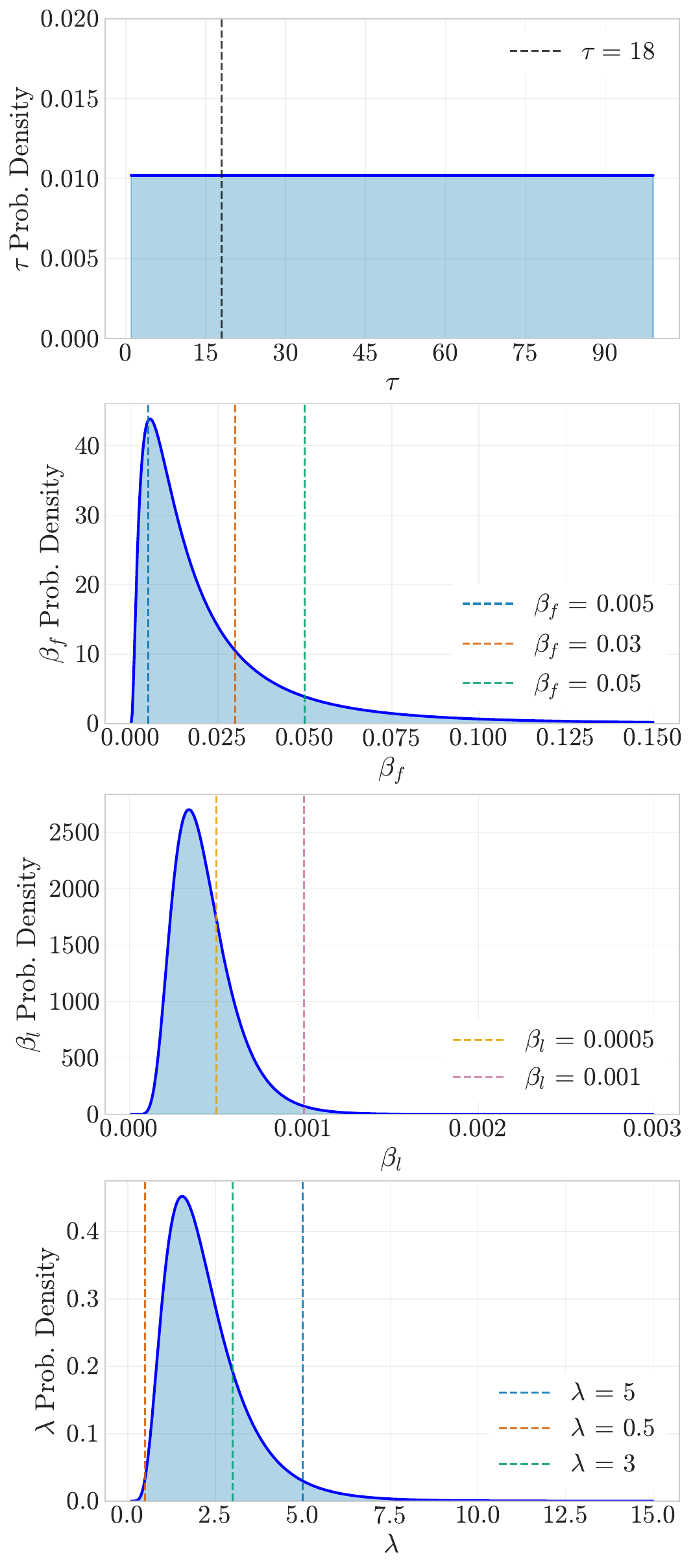}
    \caption{Prior probability densities for the changepoint time $\tau$, fouling strength $\beta_f$, leak rate $\beta_l$, and fouling-event arrival rate $\lambda$ used in the heat-exchanger model. Vertical dashed lines mark true parameter values for the study scenarios reported in Table \ref{tab:scenario_parameters}.}
    \label{fig:priors}
\end{figure}

For the operational scenario detection, a categorical prior is required. 
As heat exchangers operate normally more often than they experience failures, we assign higher prior probability to the no-failure mode: $p(\mathrm{failure\, mode})$ = [0.4; 0.2; 0.2; 0.2] \quad \text{(none, fouling, leakage, both).}
We note that the scenario where both failure mechanisms are active, is a 
very unlikely situation: albeit we still consider it in defining the categorical prior, 
we do not generate data with both leakage and fouling and consider it only as a confounding 
scenario. \\
For the changepoint ($\tau$), we assume a uniform prior over the interval $\tau \in [1, T-1]$,
excluding boundary cases in which the failure is either already present 
$(\tau \leq 0)$ or does not occur within the observation window $(\tau \geq T)$. The positive degradation parameters are assigned log-normal priors:
\begin{align*}
\beta_f &\sim \mathrm{LogNormal}(\log 0.015,\, 1.0), \\
\beta_l &\sim \mathrm{LogNormal}(\log 0.0004,\, 0.4), \\
\lambda &\sim \mathrm{LogNormal}(\log 2.0,\, 0.5).
\end{align*}
These priors enforce positivity while remaining broad enough to cover the range of fouling and leakage behaviors considered in the study as shown in Fig.~\ref{fig:priors}.

\subsection{Experimental setup}\label{sec:exp}
Process data were generated by modelling the heat exchanger in a two-stage procedure. In the first stage, we specify a parameter vector $x$ that characterizes the stationary properties of the system, such as the failure mode, failure strength, time constant $\tau$, and related quantities. Given these parameters, we generate a trajectory $z$ from a stochastic process. This trajectory represents the latent (unobserved) timeseries for fouling and leakage dynamics.
In the second stage, the latent trajectory $z$ is used as input to a simulator of the heat exchanger to generate the observable time series $y$, such as temperatures, flows, and other measured signals.\\
In its most vanilla form, simulation-based inference (SBI) is trained by providing it with query access to the simulator, yielding pairs $(x, y)$. At inference time, SBI produces an approximation to the posterior distribution $p(x \mid y)$. Since the latent trajectory $z$ is never observed and the inference procedure is not informed about the internal latent structure of the simulator SBI has no opportunity to directly infer the realized trajectory $z$. This is in contrast to MCMC methods, which have access to the full probabilistic model. A practical SBI approach is therefore to infer $x$ given $y$, and then sample trajectories $z$ from the simulator conditioned on the inferred parameters. However, if the latent variable is stochastic this procedure can never recover the actual realized trajectory, only samples from its conditional distribution as implied by the simulator. Alternatively, one could modify the inference setup and train SBI on pairs $(x, z)$ mapped to observations $y$. At inference time, conditioning on $y$ would then yield a joint posterior over both $x$ and the trajectory $z$. In principle, this allows direct estimation of latent trajectories rather than only their distributional family. In practice, however, this substantially increases the complexity of the inference problem, as SBI must explore a very high-dimensional latent space corresponding to the full time series $z$.\\
We benchmark the amortized SBI approach against a MCMC baseline. For MCMC, the stochastic degradation model is implemented in NumPyro \cite{phan2019composable, bingham2019pyro}, and the No-U-Turn Sampler \cite{NUTS} draws joint samples of the discrete failure mode $z$ and the continuous degradation parameters. This requires extensive simulator evaluations for every new observation. We found sufficient a setup made of 4 chains, each of $2,000$ warm up steps and $3,000$ samples, for a total of $20,000$ MCMC evaluations per inference task. 
For SBI, we employ the scheme depicted in Figure \ref{fig:SBI} and in Ref.\cite{papamakarios2018fastepsilonfreeinferencesimulation}. 
The neural density estimator is trained offline on a 
conservative $50,000$ forward simulations, learning a direct mapping from the 25-dimensional summary statistics of the observed parameters to an approximate posterior distribution for the failure mode and failure parameters.  For the neural density estimator within the Sequential Neural Posterior Estimation (SNPE) framework, we employed a Neural Spline Flow (NSF) architecture \cite{durkan2019neural, Greenberg2019AutomaticPT}. Unlike simpler affine autoregressive flows, NSFs utilize monotonic rational-quadratic splines, providing the high representational capacity required to map the 25-dimensional summary statistics to the complex, potentially multimodal posterior distributions of the heat exchanger's latent degradation parameters. The normalizing flow was configured with a standard multivariate normal base distribution and comprised $5$ sequential transforms. The underlying conditioner network was implemented as a multi-layer perceptron (MLP) featuring $2$ hidden layers with $50$ hidden units each, utilizing Rectified Linear Unit (ReLU) activations. The neural network was trained using the Adam optimizer with an initial learning rate of $5 \times 10^{-4}$ and a batch size of $256$. To prevent overfitting during the offline amortization phase, training was subject to an early stopping criterion, halting if the validation log-probability failed to improve over $20$ consecutive epochs. All inference architectures were built and trained utilizing the PyTorch-based \texttt{sbi toolkit} \cite{BoeltsDeistler_sbi_2025}.\\
To evaluate robustness, we use the six benchmark scenarios described in Sec.~\ref{sec:methods}. For each scenario, we evaluate $500$ independent observation records generated with the same ground-truth parameters but differing measurement noise realizations.

\section{Results}
\begin{figure}[t]
    \centering
    \includegraphics[width=1\columnwidth]{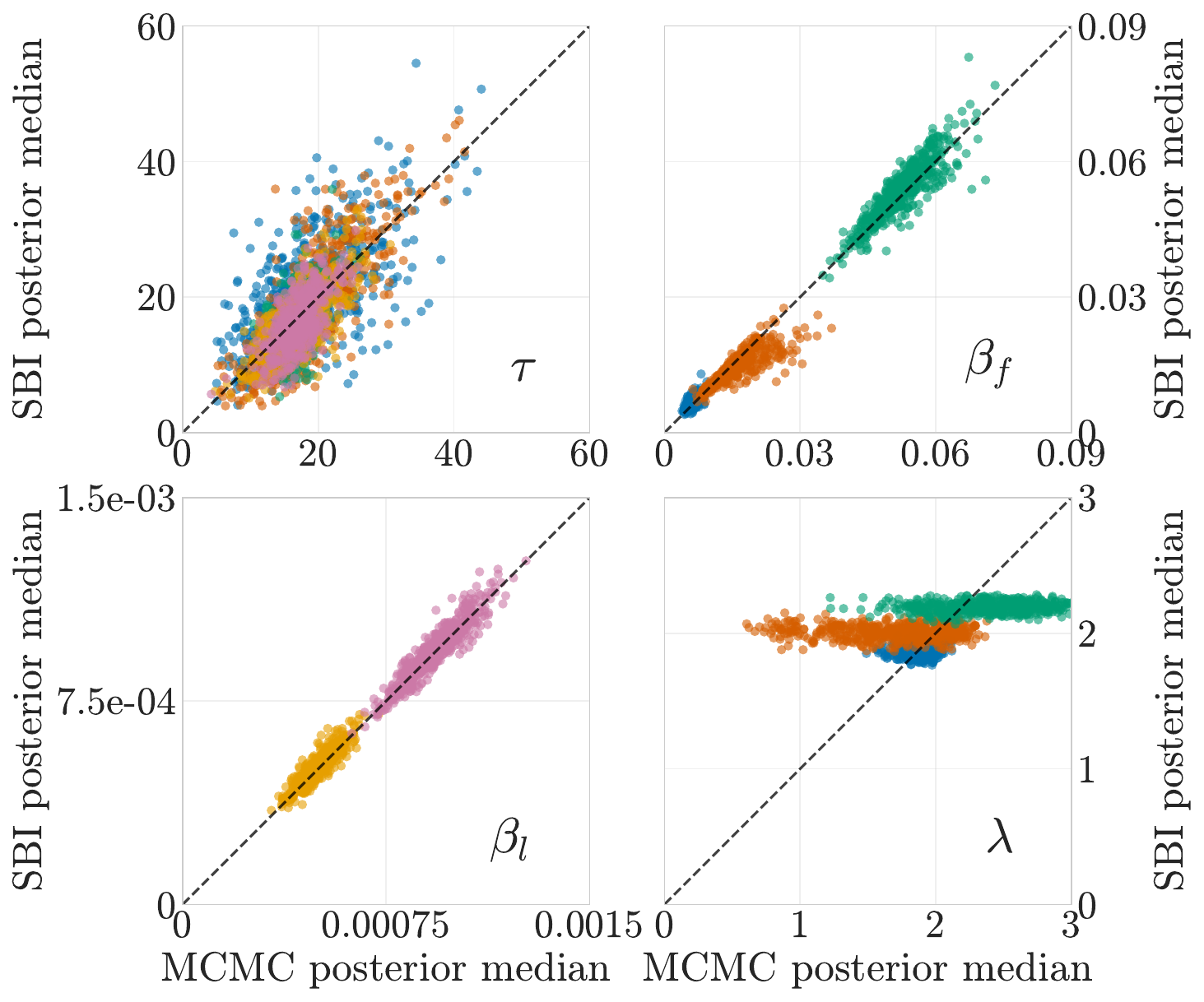}
    \caption{Scatterplot comparing posterior median parameter estimates inferred by MCMC and SBI across the 2,500 failure-case observations (500 realizations for each of Scenarios 1--5). Colors denote scenarios: blue (Weak Fouling), orange (Batch Shutdown), green (Boiler Feedwater), yellow (Mild Leakage), and lilac (Severe Leakage).
    The dashed identity line indicates perfect agreement between the two inference methods.}
    \label{fig:scatter_posterior}
\end{figure}
\begin{figure*}[t]
    \centering
    \includegraphics[width=1\textwidth]{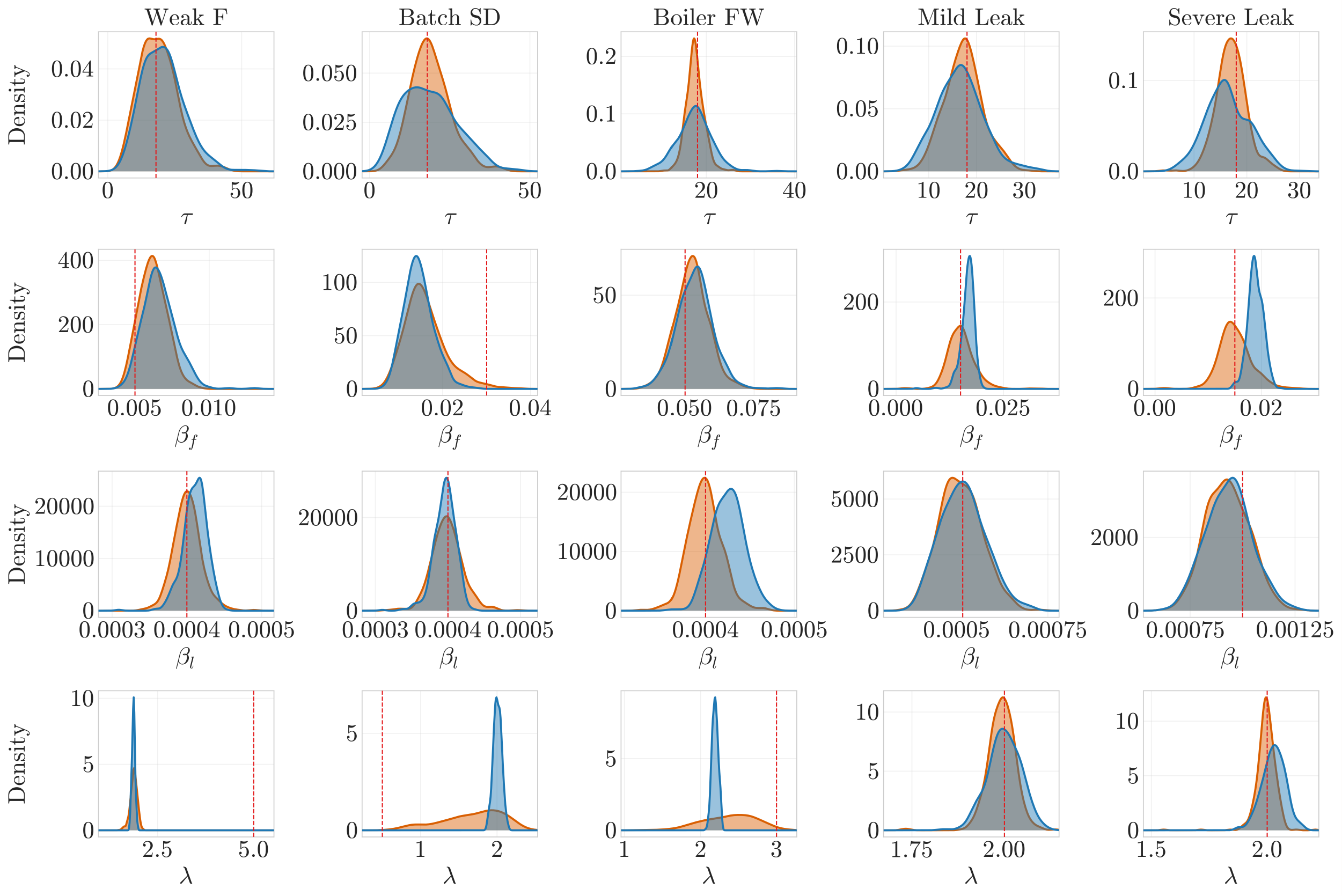}
    \caption{The marginalized distributions of the posterior medians inferred via MCMC (orange) and SBI (blue) across 500 noisy observations per scenario. The red dashed lines indicate the ground-truth parameter values. Each row represents a single continuous parameter, while each column corresponds to a specific failure scenario.}
    \label{fig:posterior_median_distributions}
\end{figure*}
This section compares SBI and MCMC over the six benchmark scenarios using 500 noisy realizations per scenario. We first assess failure-mode identification, then compare continuous-parameter inference quality and posterior shape agreement, and finally discuss Scenario 2 as a representative stress test under weak observability.
Table~\ref{tab:categorical_accuracy} shows that both MCMC and SBI identify fouling and leakage scenarios with near-perfect reliability, assigning most posterior mass to the correct failure mode across the 500 datasets per scenario. The healthy state (Scenario 6) is slightly more challenging, as transient sensor noise can emulate weak degradation and occasionally shift posterior mass toward a fault mode. Even in this regime, SBI remains statistically equivalent to the MCMC reference for failure-mode classification.
\begin{table}[t]
\centering
\small
\begin{tabular}{l  l  c  c}
\hline
\textbf{Scenario} & \textbf{Failure} & \textbf{MCMC} & \textbf{SBI} \\
\textbf{} & \textbf{mode} & \textbf{accuracy} & \textbf{accuracy} \\
\hline
1: Weak Fouling  & Fouling & 100\% & 100\% \\
2: Batch SD & Fouling & 100\% & 100\% \\
3: Boiler FW & Fouling & 100\% & 100\% \\
4: Mild Leak & Leakage & 99.8\% & 100\% \\
5: Severe Leak & Leakage & 99.6\% & 100\% \\
6: No Failure & None & 98.2\% &  98.6\% \\
\hline
\end{tabular}
\caption{Categorical failure-mode identification accuracy for MCMC and SBI across the six scenarios. Examples of time evolution curves for the scenarios are shown in Fig \ref{fig:fouling_model}.
%Accuracy drops slightly below 100\% for the healthy state (No Failure) as signal noise can occasionally resemble early-stage degradation.
}
\label{tab:categorical_accuracy}
\end{table}
Figure~\ref{fig:scatter_posterior} provides a pointwise comparison of posterior medians obtained with MCMC (horizontal axis) and SBI (vertical axis) for all failure realizations. Each marker corresponds to one inference task, so the concentration of points around the identity line indicates sample-by-sample agreement between the amortized and sampling-based estimators. Across scenarios, the cloud is tightly aligned with the diagonal for mode-discriminative parameters, confirming that SBI preserves the same central diagnostic conclusions as MCMC. The largest spread is observed for the changepoint $\tau$, particularly in sparse-event fouling regimes, where delayed and weakly informative transients reduce practical identifiability.\\
An additional pattern in Fig.~\ref{fig:scatter_posterior} is the comparatively narrow SBI posterior for $\lambda$ relative to MCMC. This behavior is consistent with an amortization-induced shrinkage effect in weakly identifiable settings: when the observation carries limited information about jump-arrival intensity, SNPE tends to regularize toward dominant simulator-supported regions learned during training, yielding sharper posteriors. In contrast, MCMC explores a flatter likelihood landscape more explicitly and therefore retains wider uncertainty for $\lambda$ \cite{delaunoy2023balancingsimulationbasedinferenceconservative, falkiewicz2023calibratingneuralsimulationbasedinference}. 
The key point is that this difference is primarily in posterior width, not in failure-mode decision; both methods still produce consistent classification outcomes.\\
\begin{figure}[t]
    \centering
    \includegraphics[width=1\columnwidth]{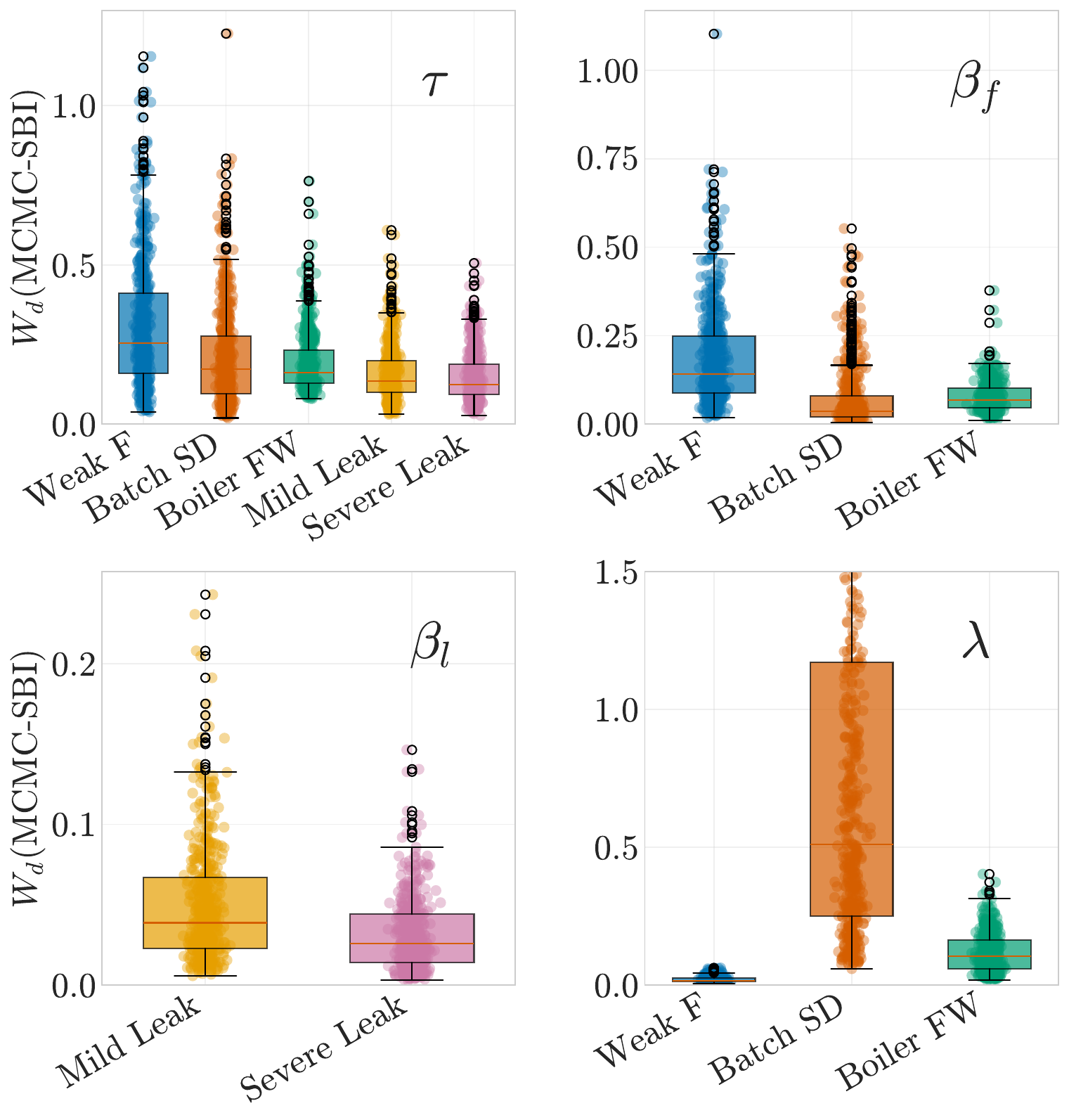}
    \caption{Normalized, 1-dimensional Wasserstein distance between MCMC and amortized SBI posterior samples for the same failure parameter, scaled by the absolute value of the true parameter. Color coding follows the one in Figure~\ref{fig:scatter_posterior}}.
    \label{fig:wasserstein}
\end{figure}
\begin{figure}[h]
    \centering
    \includegraphics[width=1\columnwidth]{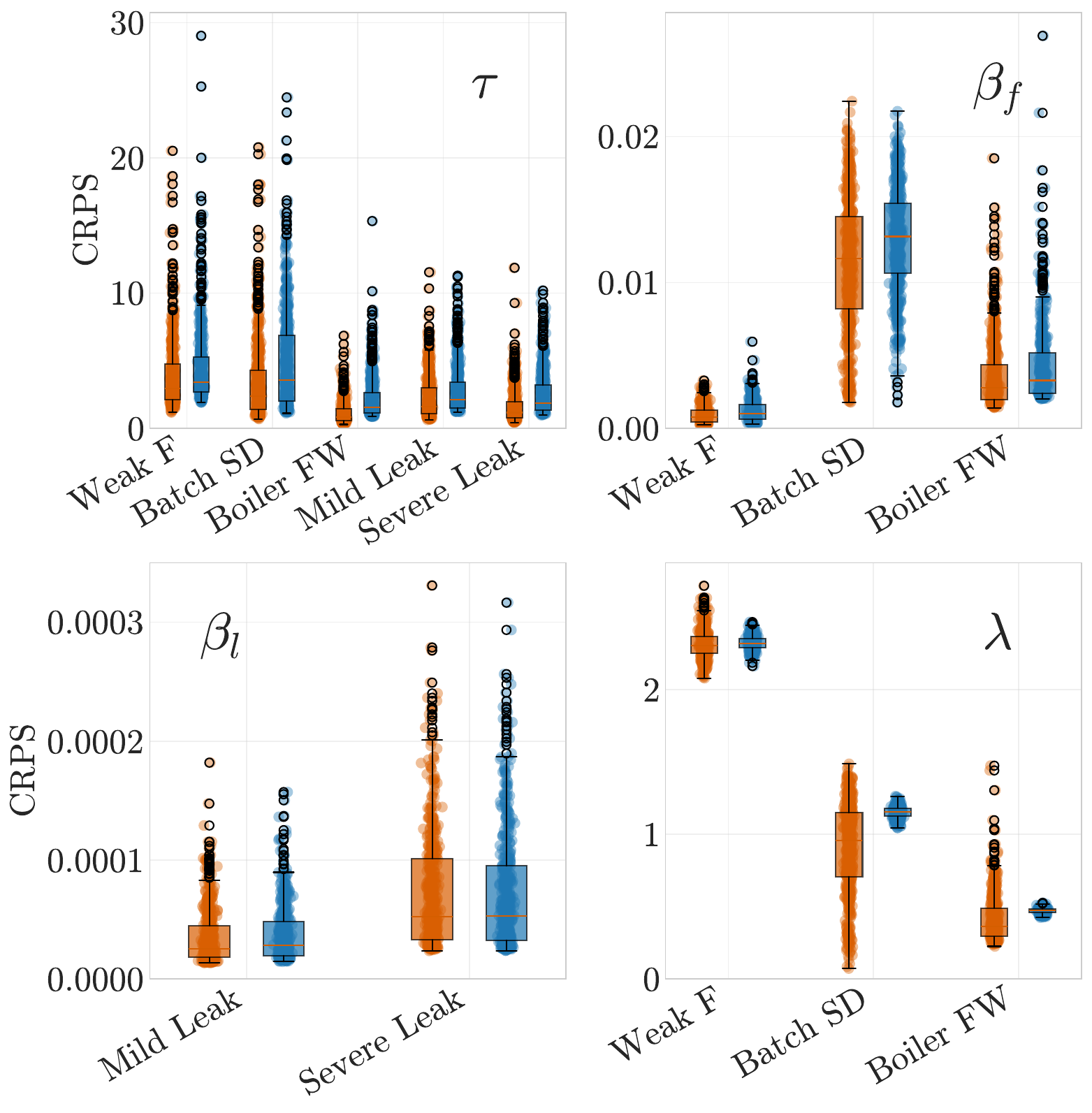}
    \caption{Continuous Ranked Probability Score (CRPS) distributions comparing the sharpness and accuracy of MCMC and SBI posterior predictions for each parameter.}
    \label{fig:crps}
\end{figure}
%\subsection{Comparing MCMC and SBI across scenarios}
To systematically compare MCMC and SBI, we analyze the inference results over the 500 independent noisy realizations per scenario. 
Fig.~\ref{fig:posterior_median_distributions} presents the marginalized distributions of posterior medians for each parameter and scenario. These distributions represent variation of point estimates across datasets, not the uncertainty width of individual posteriors. For the induction time $\tau$, medians are centered close to the ground truth in all scenarios, with MCMC yielding slightly narrower between-dataset spread. For $\beta_f$, $\beta_l$, and $\lambda$, SBI and MCMC remain broadly consistent in central tendency, while both methods show reduced accuracy for $\lambda$ in fouling scenarios. This behavior reflects identifiability limits of the stochastic degradation model rather than a method-specific failure. Consistent with Fig.~\ref{fig:scatter_posterior}, SBI posteriors for $\lambda$ are often narrower than MCMC in weakly identifiable regimes, whereas MCMC preserves broader uncertainty by exploring near-flat likelihood regions. Importantly, despite these limits in continuous-parameter recovery, failure-mode classification remains correct in virtually all $2{,}500$ failure cases. Similar qualitative conclusions are obtained when using posterior maximum-likelihood point summaries.\\
\begin{figure}%[t]
    \centering
    \includegraphics[width=0.944\columnwidth]{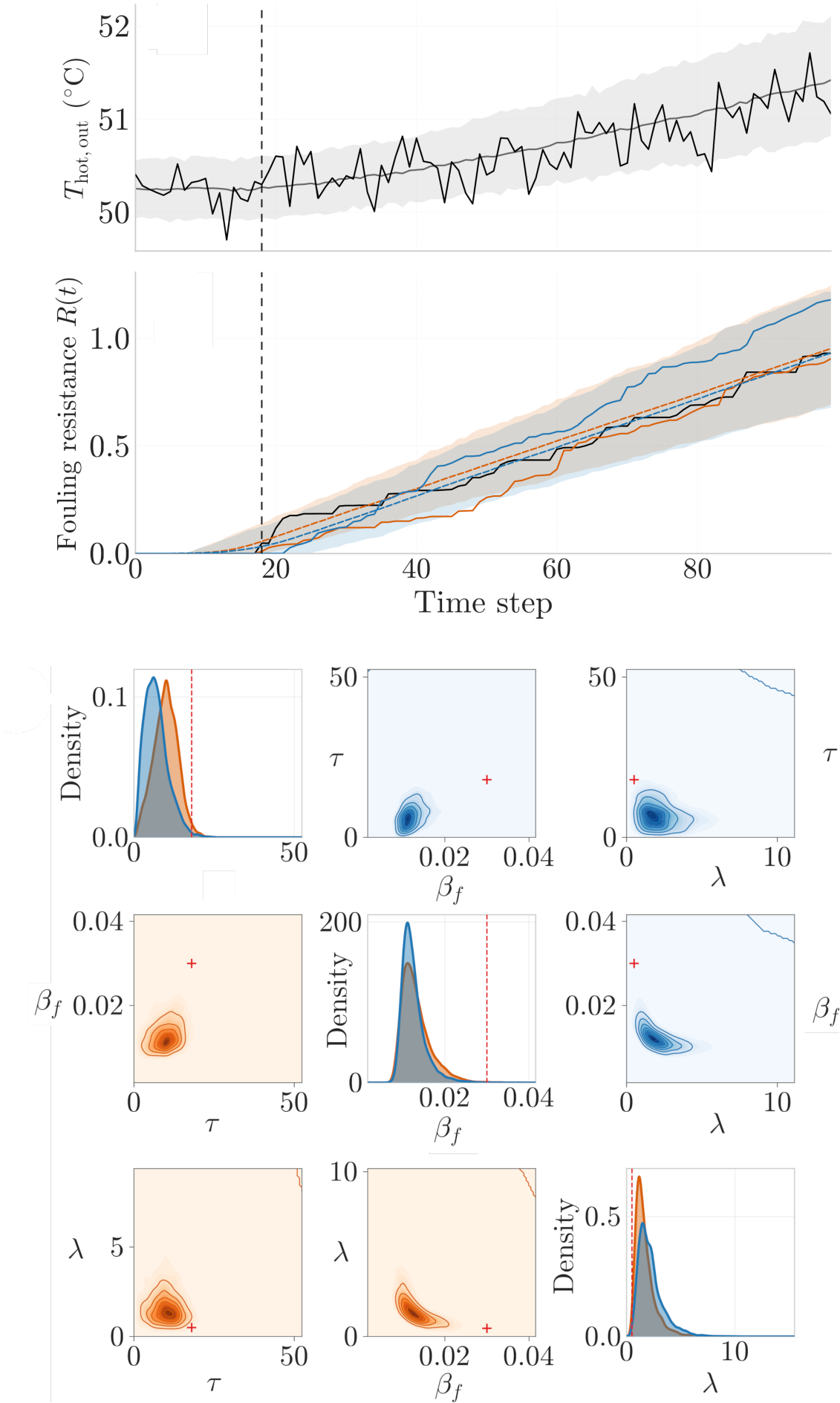}
    \caption{Posterior predictive checks for a Batch Process Shutdown (Scenario 2). 
    Top panel: generated sensor data for $T_{hot,out}$ (full line), average over 500 
    realizations (dashed line) and 90\% confidence interval. Vertical dashed line show failure onset time ($\tau$). 
    Mid panel: Fouling resistance 
    computed for the observed realization (black) together with an example trajectory from parameters sampled from the posteriors inferred by MCMC (orange) and SBI (blue). The average (dashed lines) and 95\% confidence intervals (shaded) are shown for trajectories generated from 500 posterior samples.
   Bottom panel: The inferred posterior probabilities per process parameter from MCMC (orange) and SBI (blue). The ground truth is shown as red lines in the diagonal density plots and crosses in the covariance plots.}
    \label{fig:posterior_s2v}
\end{figure}
To assess similarity of full posterior shapes, we evaluate the 1D Wasserstein distance between MCMC and SBI posteriors for each parameter (Fig.~\ref{fig:wasserstein}). The distributions are generally concentrated at low values, indicating that the amortized SBI posterior remains close to the MCMC reference over most inference tasks. Larger distances appear primarily in weakly identifiable regimes, where multiple parameter combinations explain the observations nearly equally well. This pattern is consistent with the scenario-level analysis above: discrepancies are driven more by structural identifiability limits than by systematic bias of the SBI estimator.\\
Finally, Fig.~\ref{fig:crps} reports the Continuous Ranked Probability Score (CRPS), a proper scoring rule that jointly evaluates probabilistic sharpness and statistical accuracy. The CRPS distributions for SBI and MCMC are of similar magnitude across parameters, indicating that SBI preserves most of the probabilistic predictive quality of the MCMC baseline.
For the induction time $\tau$ and degradation parameters $\beta_f$, $\beta_l$, both metrics demonstrate acceptable predictive accuracy levels in most scenarios.\\
The exception is the determination of $\lambda$ in the Batch Process Shutdown (Scenario 2), where the posterior location is consistently overestimated due to the extreme sparsity of the fouling jumps. In this regime, the underlying degradation is driven by a very low event arrival rate ($\lambda = 0.5$) combined with a large jump scale ($\beta_f = 0.03$). Because these stochastic shocks are exceedingly rare, a given finite observation window may capture very few—or even zero—fouling events. Consequently, the observed thermal-fluid signals carry minimal information regarding the true arrival rate, resulting in a nearly flat likelihood surface for $\lambda$. In the absence of informative data, the posterior distribution naturally shrinks toward the prior. Since the assigned log-normal prior for $\lambda$ is centered at a median of $2.0$ —significantly higher than the true scenario value—both MCMC and SBI predictably overestimate this parameter. This behavior highlights a fundamental limit of structural identifiability in sparse-event regimes, rather than a systematic failure of the inference algorithms.\\
The challenge is compounded by a fundamental identifiability trade-off between $\lambda$ and $\beta_f$. Infrequent large jumps and frequent small jumps can produce near-identical cumulative fouling trajectories, making it difficult to distinguish the true regime from the dominant prior. MCMC is particularly susceptible because it must simultaneously infer $\lambda$ alongside 100 hidden per-timestep jump indicators, entangling the global arrival rate with local decisions about whether each individual timestep carried a jump. SBI avoids this by absorbing the individual timestep variables into the simulation process and instead relies on summary statistics. This compresses what little signal exists about the rate of fouling accumulation into a lower-dimensional representation. Even so, neither method can fully overcome the prior dominance when the observation window contains so few events. This highlights a fundamental limit of structural identifiability in sparse-event regimes, rather than a systematic failure of the inference algorithms.
Detailed model calibration analyses are provided in the Supplementary Material.\\
Scenario 2 (Batch Process Shutdown) provides therefore a representative stress test for probabilistic diagnosis under weak observability. 
In this regime, fouling evolves through sparse but potentially large stochastic jumps (low event rate, $\lambda=0.5$, and relatively
large jump scale, $\beta_f=0.03$), so the thermal response can resemble measurement noise over extended time windows. This setting is
therefore informative for assessing whether the inference framework can separate persistent degradation dynamics from transient fluctuations.
Both MCMC and SBI assign dominant posterior probability to the fouling mode (Table \ref{tab:categorical_accuracy}) and recover
a consistent posterior for the changepoint $\tau$. 
As shown in Fig.~\ref{fig:posterior_s2v}, marginal posteriors are also in close agreement across methods, indicating
that amortized SBI preserves the main uncertainty structure of the MCMC reference posterior. Although both approaches show limited
identifiability for $\lambda$ and, to a lesser extent, $\beta_f$ in this sparse-event regime, posterior predictive trajectories
remain physically plausible and reproduce the observed trend in fouling resistance. From a reliability perspective, this is the 
key requirement for downstream condition monitoring tasks, such as predicting the Remaining Useful Life of the equipment.
Even when individual parameters are weakly identifiable, the inferred posterior 
ensemble still supports robust fault recognition and uncertainty-aware forecasting of degradation progression.
Accurately recovering these parameters and their associated uncertainties is a critical first step for downstream

\subsection{Resources consumption}
\begin{figure}[t]
    \centering
    \includegraphics[width=0.85\columnwidth]{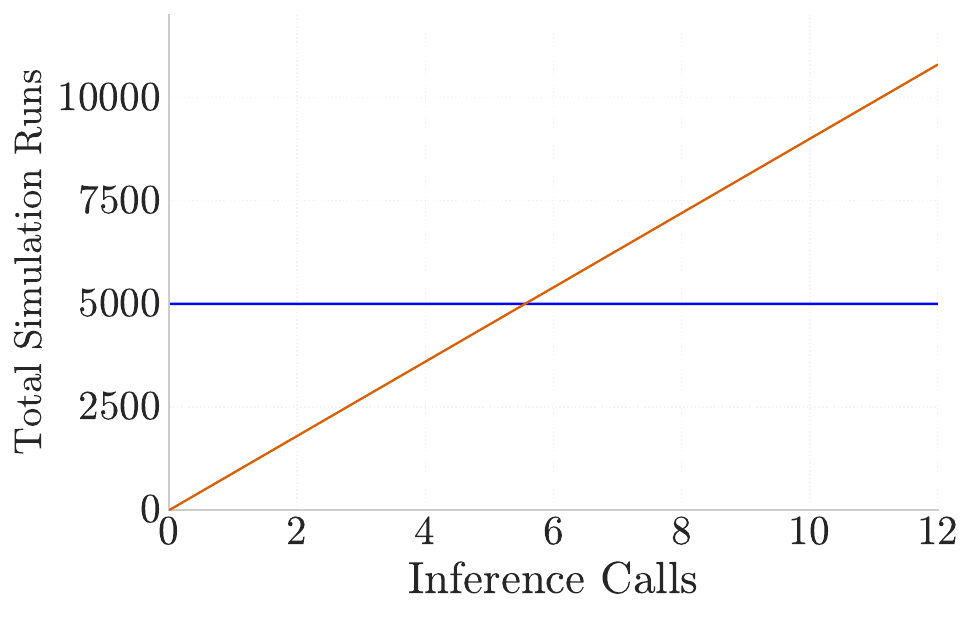}
    \caption{%
  Resource comparison at the minimal configuration achieving both
  ${\geq}95\%$ classification accuracy on Scenario~2
  (fouling, $\tau=18$).
  MCMC used 4 chains with 150 burn-in $+$ 75 samples (900 transitions/call);
  SBI used 5{,}000 training simulations.
  SBI is 82$\times$ faster per call after a break-even of $\sim$6 inference calls.}
    \label{fig:SBI_MCMC_results}
\end{figure}

\begin{table}[b]
\centering
\small
\begin{tabular}{lcccc}
\hline
\textbf{Method} & \textbf{Sim.\ calls} &\textbf{ Acc.} & \textbf{Infer.\ time} & \textbf{Train cost} \\
\hline
MCMC & 900 / call   & 100\,\% & $2.4$\,s / call          & ---   \\
SBI  & 5{,}000 (once) & 100\,\% & $0.029$ s / call & 19\,s \\
\hline
\end{tabular}
\caption{Computational cost of inference at 100$\%$ accuracy for MCMC and SBI on an Apple M4 Pro laptop.}
\label{tab:resource_comparison}
\end{table}
The primary motivation for adopting SBI in industrial asset management is its exceptional computational scalability at deployment time. As summarized in Table~\ref{tab:resource_comparison}, MCMC must restart its iterative sampling process from scratch for every new diagnostic request (inference), so its total simulation cost grows linearly with the number of inference tasks.\\
In contrast, SBI training is a \emph{one-time} offline phase whose upfront cost is rapidly amortized over subsequent diagnoses. 
We therefore pushed both approaches to a compact setup where the simulator calls are as few as possible while 
retaining acceptable accuracy on the parameters posteriors, in order to test 
the computational cost of inference in the two cases. MCMC was run on 4 chains with 150 burn-in and 75 samples (900 transitions/call) while SBI was trained on 5,000 simulations.
As illustrated in Fig.~\ref{fig:SBI_MCMC_results}, the two methods reach cost parity after fewer than six inference calls, beyond which SBI is 82 times faster per diagnosis while maintaining identical classification accuracy and $\tau$ convergence.\\
In this study, the simulator is a lightweight JAX-JIT-compiled effectiveness-NTU model of a single heat exchanger, chosen as a toy model
for testing. Each simulator call is estimated to require 0.03 $s$ of computing on a 4-core CPU Apple M4 Pro with JAX-JIT pre-warmed. 
Therefore, the absolute time saved per call is modest.
However, the 82$\times$ acceleration is a property of the inference algorithms rather than of the particular simulator, and it scales with simulator cost. For networks of interacting heat exchangers or plant-wide process simulations where a single forward evaluation can take minutes to hours, the same factor translates to savings of tens of minutes to many hours \emph{per diagnosis}. Across a modern process plant where condition monitoring must be performed simultaneously on dozens of assets at high frequency, the near instantaneous posterior evaluation offered by SBI provides a pathway to real-time, risk-aware decision-making that is entirely intractable with classical Bayesian sampling.

\section{Discussion}
This study demonstrates that simulation-based inference (SBI) provides a practical and scalable alternative to classical Bayesian sampling for condition monitoring of heat exchangers. Across all evaluated scenarios, SBI achieves diagnostic performance comparable to MCMC in both failure-mode classification and parameter estimation, while reducing inference time by several orders of magnitude. The close agreement in posterior summaries and distributional metrics indicates that SBI with amortized neural posterior estimation can efficiently approximate high-fidelity MCMC baselines, even in nonlinear thermo-fluid systems with stochastic degradation dynamics.\\
A key implication is the shift in computational burden enabled by SBI. By front-loading simulation cost into an offline training phase, the framework enables near-instantaneous inference during deployment. This is particularly relevant for industrial settings where continuous monitoring must be performed across multiple assets under strict latency constraints. The results suggest that, beyond a small number of inference queries, SBI becomes more computationally efficient than MCMC, making it suitable for real-time diagnostics and large-scale deployment.\\
From a reliability engineering perspective, the framework supports probabilistic fault diagnosis by providing full posterior distributions over failure modes and degradation parameters. 
This enables uncertainty-aware decision-making, which is critical for risk assessment, maintenance planning, and integrating uncertainty information directly into robust RUL predictions \cite{XU2024110250}. Notably, the results indicate that failure onset and mode identification is highly robust, even in regimes where individual parameters are weakly identifiable. This distinction highlights that reliable anomaly detection may be achievable even when precise quantification of degradation dynamics remains challenging.\\
Several limitations should be considered. First, parameter identifiability depends strongly on the observability of the underlying process; for example, the fouling intensity and changepoint parameters are difficult to estimate in scenarios with sparse or inactive degradation. These challenges arise from both measurement noise and structural properties of the stochastic model. Second, the study relies on synthetic data generated from a simplified degradation model, which does not fully capture the complexity of real industrial systems. In particular, assumptions such as stationary noise and simplified fouling dynamics may limit direct transferability. Additionally, the use of engineered summary statistics, while computationally efficient, may discard information present in the full time series.\\
Despite these limitations, the approach is broadly applicable due to its model-agnostic nature. SBI does not require an explicit likelihood function and can be applied to black-box simulators, making it suitable for legacy systems where governing equations are inaccessible. Compared to purely data-driven methods, the framework retains physical interpretability through latent parameters while providing calibrated uncertainty estimates.\\
Future work should focus on validation with real operational data, where additional sources of variability such as sensor drift and unmodeled disturbances are present. However, for real-life industrial systems of increased complexity, the MCMC baseline will be computational infeasible and hence the verification will most likely be limited to few hand-labeled failures. Extensions to adaptive or online training could improve robustness under distributional shift, while joint inference of latent trajectories and parameters may further enhance diagnostic resolution. More generally, integrating SBI-based inference into digital twin frameworks offers a pathway toward scalable, probabilistic condition monitoring in complex industrial systems.

\section{Conclusions}
This work introduces a Bayesian framework for heat exchanger condition monitoring using simulation-based inference (SBI) powered by amortized neural posterior estimation.
The results show that SBI matches MCMC in diagnostic accuracy while reducing inference time by a factor of 82.
This massive computational acceleration enables near real-time estimation of failure modes and latent degradation parameters.
The method remains robust across a range of scenarios, including sparse-event cases with weak parameter identifiability.
From a reliability perspective, the framework provides the uncertainty-aware diagnostics essential for risk-informed decision-making and predictive maintenance.
Crucially, its amortized and model-agnostic nature allows for the continuous inference of process and health parameters in realistic, plant-wide industrial settings where traditional MCMC sampling is entirely computationally unfeasible.
Future work will focus on validation with real operational data and improving robustness under model mismatch.

\section{Data and code availability}
The code and experiments used in this paper are available on GitHub
at: \nolinkurl{https://github.com/petercollett-cognite/sbi_mcmc_heat_exchanger.git}.

%\section{CRediT authorship contribution statement}
\section{authorship contribution statement}
\textbf{Peter Collet}:  Writing – review $\&$ editing, formal analysis.
\textbf{Simone Casolo}: Writing – review $\&$ editing, formal analysis, conceptualization.
\textbf{Alexander J. Stasik}: Review $\&$ editing, supervision, methodology, conceptualization.
\textbf{Signe Riemer-S\o rensen}: Writing – review $\&$ editing, supervision, methodology, conceptualization.\\

\section{Declaration of competing interest}
The authors declare that they have no known competing financial interests or personal relationships that could have appeared to influence the work reported in this paper.

\section{Acknowledgements}
This publication has been funded by the SFI NorwAI, (Centre for Research-based Innovation, 309834). The authors gratefully acknowledge the financial support from the Research Council of Norway and the partners of the SFI NorwAI.

\bibliographystyle{apsrev4-2}
\bibliography{references}

\end{document}